\documentclass{article}
                  \usepackage{hyperref}
                  \usepackage{natbib}
                  \usepackage{amssymb}
                  \usepackage{authblk}

\usepackage{bbm}
\newcommand{\indic}[1]{\mathbbm{1}\left[#1\right]}
\usepackage{caption}
\usepackage{booktabs}
\usepackage[table]{xcolor}
\usepackage{mathtools}
\DeclareMathOperator{\E}{\mathbb{E}}
\DeclarePairedDelimiter{\floor}{\lfloor}{\rfloor}
\newcommand*\dif{\mathop{}\!\mathrm{d}}
\author{Taylor Pospisil}
\author{Ann B. Lee}
\affil{Department of Statistics \& Data Science \\ Carnegie Mellon University \\ Pittsburgh, PA 15289, USA}
\newcommand{\acks}[1]{\section*{Acknowledgements} #1}
\date{}
\title{RFCDE: Random Forests for Conditional Density Estimation}
\begin{document}

\maketitle
\begin{abstract}
Random forests is a common non-parametric regression technique which
performs well for mixed-type data and irrelevant covariates, while
being robust to monotonic variable transformations. Existing random
forest implementations target regression or classification. We
introduce the \texttt{RFCDE} package for fitting random forest models
optimized for nonparametric conditional density estimation,
including joint densities for multiple responses. This enables
analysis of conditional probability distributions which is useful
for propagating uncertainty and of joint distributions that describe
relationships between multiple responses and covariates. \texttt{RFCDE} is
released under the MIT open-source license and can be accessed at
\url{https://github.com/tpospisi/rfcde}. Both R and Python versions, which
call a common C++ library, are available.
\end{abstract}

\section{Introduction and Motivation}
\label{sec:orgd689b66}
Conditional density estimation (CDE) is the estimation of the
density \(f(z \mid X = x)\) where we condition the response \(Z\) on
observed covariates \(X\). In a prediction context, CDE provides a
more nuanced accounting of uncertainty than a point estimate or
prediction interval, especially in the presence of heteroskedastic
or multimodal responses. These conditional densities can be used to
propagate uncertainty through further analyses or to minimize
expected prediction loss for non-standard loss functions.

Our main contribution in \texttt{RFCDE} is two-fold: (1) we provide
software for extending random forest estimation to conditional
density estimation including joint densities for multivariate
responses, and (2) we fit our trees based upon minimizing
conditional density estimation loss. This overcomes the limitations
of the usual regression loss functions due to heteroskedasticity and
multimodality while still remaining computationally feasible.

We extend random forests \citep{breiman2001random} to CDE, inheriting
the benefits of random forests with respect to mixed-data types,
irrelevant covariates, and data transformations. We take advantage
of the fact that random forests can be viewed as a form of adaptive
nearest-neighbor method with the aggregated tree structures
determining a weighting scheme. This weighting scheme can then be
used to estimate quantities other than the conditional mean; in this
case the conditional density.

Other existing random forest implementations such as
\texttt{quantileregressionForests} \citep{meinshausen2006quantile} and \texttt{trtf}
\citep{hothorn2017transformation} can be used for CDE. In Section
\ref{orga5c1efe} we show that our method achieves lower CDE loss in
competitive computational time.

For ease of use in the statistics and machine-learning communities,
we provide packages in both R and Python. Both packages call a
common C++ library which implements the core training functions for
performance and avoidance of redundancies. The core library can
easily be wrapped in other languages. Source code and a bug tracker
can be found at \url{https::/github.com/tpospisi/rfcde}.

\section{Overview}
\label{sec:org2d34285}
To construct our estimators we largely follow the usual random
forest construction. At their simplest, random forests are
ensembles of regression trees. Each tree is trained on a
bootstrapped sample of the data. The training process
involves recursively partitioning the covariate space through
splitting rules taking the form of splitting into the sets
\(\left\{X_{i} \le v\right\}\) and \(\left\{X_{i}> v\right\}\) for a
particular covariate \(X_{i}\) and split point \(v\). Once a partition
becomes small enough (controlled by the tuning parameter
\texttt{nodesize}), it becomes a leaf node and is no longer partitioned.

For prediction we use the tree structure to calculate weights for
the training data from which we perform a \emph{weighted kernel density
estimate} using ``nearby'' points. Borrowing the notation of
\cite{breiman2001random} and \cite{meinshausen2006quantile}, let \(\theta_{t}\)
denote the tree structure for a single tree. Let \(R(x, \theta_{t})\) denote
the region of covariate space covered by the leaf node for input
\(x\). Then for a new observation \(x^{*}\) we use \(t\)-th tree to
calculate weights for each training point \(x_{i}\) as

\begin{equation*}
w_{i}(x^{*}, \theta_{t}) = \frac{\indic{X_{i} \in R(x^{*}, \theta_{t})}}{\sum_{i=1}^{n} \indic{X_{i} \in R(x^{*}, \theta_{t})}}.
\end{equation*}

We then aggregate over trees setting \(w_{i}(x^{*}) = T^{-1}\sum_{t=1}^{T}
   w_{i}(x^{*}, \theta_{t})\). The weights are finally used for the weighted
kernel density estimate.

\begin{equation}
\label{eq:orgb8815ef}
  \widehat{f}(z \mid x^{*}) = \frac{1}{\sum_{i=1}^{n} w_{i}(x^{*})} \sum_{i=1}^{n} w_{i}(x^{*}) K_{h}(Z_{i} - z)
\end{equation}

where \(K_{h}\) is a kernel function integrating to one.

Our main departure from the standard random forest algorithm is
choosing the splits of the partitioning. In regression contexts,
the splitting variable and split point are often chosen to minimize
the mean-squared error loss. For CDE, we instead choose to minimize
a \emph{loss specific to CDE} \citep{izbicki2017converting}

\begin{equation*}
  L(f, \widehat{f}) = \int \int \left(f(z \mid x) - \widehat{f}(z \mid x)\right)^{2} \dif z \dif P(x).
\end{equation*}

This loss is the \(L^{2}\) error for density estimation weighted by the
marginal density of the covariates. To conveniently estimate this
loss we can expand the square and rewrite the loss as

\begin{equation}
\label{eq:org40e65dc}
  L(f, \widehat{f}) = \E_{X}\left[\int \widehat{f}^{2}(z \mid X) \dif z\right] - 2 \E_{X, Z}\left[\widehat{f}(Z \mid X)\right] + C_{f}
\end{equation}

with \(C_{f}\) as a constant which doesn't depend on \(\widehat{f}\). The
first expectation is with respect to the marginal distribution of
\(X\) and the second with respect to the joint distribution of \(X\)
and \(Z\). We estimate these expectations by their empirical
expectation on observed data.
While we use kernel density estimates for predictions on new
observations, we do not use kernel density estimates when
evaluating splits because of computationally expensive calculations
in Equation \ref{eq:org40e65dc} due to the dependence of
\(\widehat{f}\) on the \(O(n^{2})\) pairwise-distances between all
training points.

For fast computations, we instead use \emph{orthogonal series} to compute
density estimates for splitting. Given an orthogonal basis \(\left\{
   \Phi_{j}(z) \right\}\) such as a cosine basis or wavelet basis, we can
express the density as \(\widehat{f}(z \mid x) = \sum_{j}
   \widehat{\beta_{j}}\phi_{j}(z)\) where \(\widehat{\beta_{j}} = \frac{1}{n_{}}
   \sum_{i=1}^{n} \phi_{j}(z_{i})\). This choice is motivated by a convenient formula
for the CDE loss associated with an orthogonal series density
estimate

\begin{equation*}
  \widehat{L}(f, \widehat{f}) - C_{f} = -\sum_{j} \widehat{\beta_{j}}^{2}.
\end{equation*}

The above expression only depends upon the quantities \(\left\{
    \widehat{\beta_{j}} \right\}\) that themselves depend only upon \emph{linear} sums
of \(\phi_{j}(z_{i})\). This makes it computationally efficient to evaluate
the CDE loss for each split.

\section{RFCDE}
\label{sec:org869e14a}
We provide \texttt{RFCDE} packages for R and Python. To avoid redundancy
and achieve better performance, the main model fitting code is
written as a common C++ library. We use \texttt{Rcpp}
\citep{eddelbuettel2011rcpp} and \texttt{Cython} \citep{behnel2011cython} to
write wrappers for R and Python. Vignettes are included on Github
for each language.

\subsection{Usage}
\label{sec:orge243da7}
The R and Python packages expose three main functions:

\begin{itemize}
\item \texttt{RFCDE(x\_train, z\_train, ...)} trains the random forest and
returns a wrapped C++ object. There are several tuning
parameters; for details see the documentation on Github.
\item \texttt{predict(x\_new, z\_grid, bandwidth)} uses the fitted forest to
estimate the weighted kernel density estimate for all points in
\texttt{z\_grid}.
\item \texttt{weights(forest, x\_new)} which uses the fitted forest to
calculate weights for each new observation. This is used
internally in \texttt{predict} but could be adapted for other purposes.
\end{itemize}

\subsection{Univariate Experiment}
\label{sec:org12c71a1}
To illustrate the difference that training the random forest to
minimize a CDE loss can have on performance, we will compare
against an existing random forest implementation (that minimizes
MSE loss). We adapt the \texttt{quantregForest}
\citep{meinshausen2006quantile} package in R to perform conditional
density estimation according to Equation \ref{eq:orgb8815ef}. This is
equivalent to our method except that the splits for
\texttt{quantregForest} minimize mean-squared error rather than CDE loss.
This provides a useful comparison to demonstrate the advantages of
specifically training random forests for the goal of conditional
density estimation. We also compare against the \texttt{trtf} package
\citep{hothorn2017transformation} which trains forests for CDE using
flexible parametric families.
\label{org2aae8ee}

We generate data from the following model

\vspace{-5mm}
\begin{eqnarray*}
  X_{1:10}, Y_{1:10} \sim \operatorname{Uniform}(0, 1), \quad S \sim \operatorname{Multinomial}(2, \frac{1}{2}) \\
  Z \mid X, Y, S \sim \begin{cases}
    \operatorname{Normal}(\floor*{\sum_{i} X_{i}}, \sigma) & S = 1 \\
    \operatorname{Normal}(-\floor*{\sum_{i} X_{i}}, \sigma) & S = 2 \\
  \end{cases}
\end{eqnarray*}

where the \(X_{i}\) are the relevant covariates with the \(Y_{i}\) serving
as irrelevant covariates. \(S\) is an unobserved covariate which
induces multimodality in the conditional densities.

Under the true model \(\E[Z \mid X, Y] = 0\), so there is no
partitioning scheme that can reduce the MSE loss (as the
conditional mean is always zero). As such, the trained
\texttt{quantregForest} trees behave similarly to nearest neighbors as
splits are effectively chosen at random.
We evaluate the two methods on two criterion: training time and
CDE loss on a validation set. All models are tuned with the same
forest parameters (\texttt{mtry} = 4, \texttt{ntrees} = 1000). \texttt{RFCDE} has
\texttt{n\_basis} = 15. \texttt{trtf} is fit using a Bernstein basis of order 5.
\texttt{RFCDE} and \texttt{quantregForest} have bandwidth 0.2.

\begin{table}[htbp]
\caption{\label{tab:org04fdf3c}
\small Performance of \texttt{RFCDE} and competing methods; smaller CDE loss implies better estimates.}
\centering
\begin{tabular}{rcccc}
Method & N & CDE Loss (SE) & Train Time (seconds) & Predict Time (seconds)\\
\hline
\texttt{RFCDE} & 1,000 & \textbf{-0.171} (0.004) & 2.31 & 1.29\\
\texttt{quantregForest} & 1,000 & -0.152 (0.003) & \textbf{1.17} & \textbf{0.61}\\
\texttt{trtf} & 1,000 & -0.109 (0.001) & 1013.54 & 123.47\\
\hline
\texttt{RFCDE} & 10,000 & \textbf{-0.194} (0.003) & \textbf{42.99} & 1.95\\
\texttt{quantregForest} & 10,000 & -0.159 (0.003) & 48.60 & \textbf{0.47}\\
\hline
\texttt{RFCDE} & 100,000 & \textbf{-0.227} (0.004) & \textbf{904.76} & 8.05\\
\texttt{quantregForest} & 100,000 & -0.173 (0.003) & 1289.93 & \textbf{0.61}\\
\end{tabular}
\end{table}

Table \ref{tab:org04fdf3c} summarizes the results for three simulations
of size 1000, 10000, and 100000 training observations. We use 1000
observations for the test set and calculate the CDE loss. We see
that \texttt{RFCDE} performs substantially better on CDE loss. We also
note that \texttt{RFCDE} has competitive training time especially for
larger data sets. \texttt{trtf} is only run for the smallest data set due
to its execution time.

\subsection{Joint Experiment}
\label{sec:org78f9690}
\label{orga5c1efe}

Another feature of \texttt{RFCDE} is the ability to target joint
conditional density loss. The splitting process extends
straightforwardly to the multivariate case through the use of a
tensor basis. The density estimation similarly extends through the
use of multivariate kernel density estimation. This allows for
conditional density estimation in multiple dimensions (although in
practice only two or three dimensions are computationally practical
due to the limitations of the tensor basis).
To showcase our model, we draw 10000 samples from the following
distribution:

\begin{equation*}
X \sim \operatorname{Uniform}(0, 1) \quad Z_{1} \sim \operatorname{Uniform}(0, X) \quad Z_{2} \sim \operatorname{Uniform}(X, Z).
\end{equation*}
We fit our model with parameters \texttt{ntrees} = 1000, \texttt{mtry} = 1,
\texttt{nodesize} = 20, and \texttt{n\_basis} = 15. The bandwidth is selected
adaptively with respect to \(x^{*}\) for each density.

\begin{figure}[htbp]
\centering
\includegraphics[width=1.0\textwidth,height=0.25\textwidth]{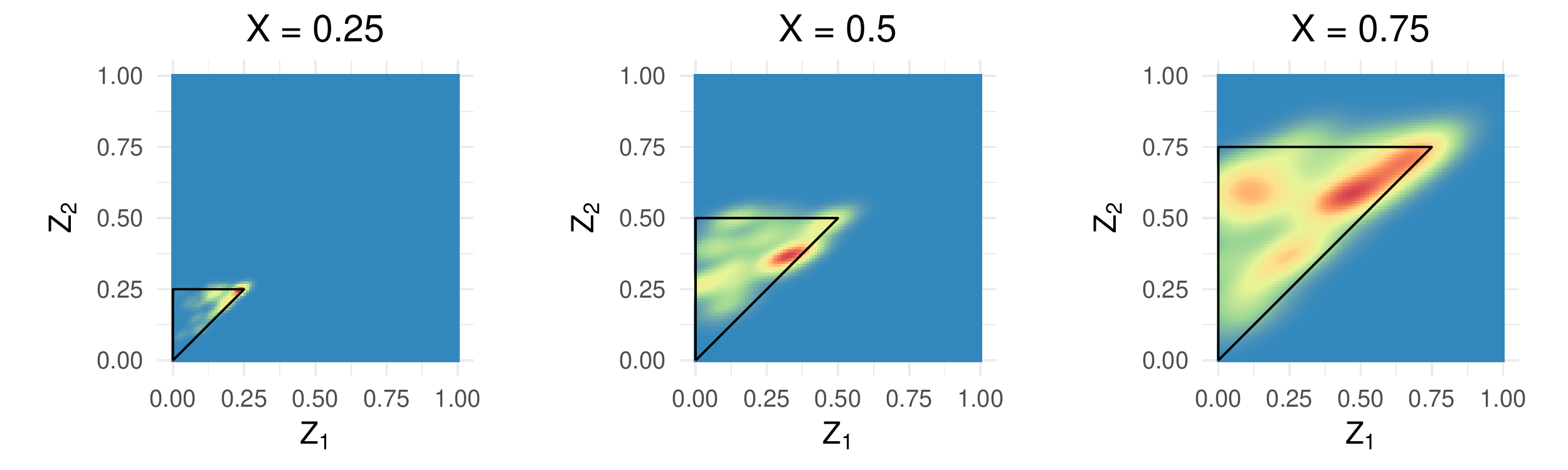}
\caption{\label{fig:org5f0d88f}
\small Joint conditional density estimates; the test values of \(X\) are listed above each joint density. We see that our method captures the joint dependence between the two responses as shown by the black lines which indicate the true support of the joint conditional density.}
\end{figure}

We cannot straightforwardly adapt \texttt{quantregForest} or \texttt{trtf}
software to joint densities so we will not compare against them.
Instead we provide a qualitative assessment of performance in
Figure \ref{fig:org5f0d88f}. The joint densities, and in
particular the dependency between \(Z_{1}\) and \(Z_{2}\), are captured
well.

\section{Conclusions}
\label{sec:org5f6dd5e}
We provide \texttt{RFCDE}, a multi-language tool for fitting conditional
density estimation (CDE) random forests for complex data and more
than one response variable. \texttt{RFCDE} extends random forests by
implementing a computationally efficient method for minimizing a CDE
loss in each split. R and Python packages are available along with a
common C++ library at \url{https://github.com/tpospisi/rfcde}.

\acks{We are grateful to Rafael Izbicki for helpful discussions and
comments on the paper. This work was partially supported by NSF
DMS-1520786.}

\bibliography{rfcde-paper}
\end{document}